\documentclass[runningheads]{llncs}
\usepackage{graphicx}
\usepackage{rotating}
\usepackage{multirow}
\usepackage{newfloat,footmisc}
\usepackage{listings}
\usepackage[hyphens]{url}
\usepackage{times}
\usepackage{epsfig}
\usepackage{graphicx}
\usepackage{amsmath}
\usepackage{amssymb}
\usepackage{color}
\usepackage{booktabs}
\usepackage{caption}
\usepackage{pifont}
\newcommand{\cmark}{\ding{51}}%
\newcommand{\xmark}{\ding{55}}%
\definecolor{Gray}{gray}{0.85} 
\definecolor{SkyBlue}{rgb}{0.88,1,1}
\usepackage[table,dvipsnames]{xcolor}
\newcommand{\spacing}{4pt}
\renewcommand\textfloatsep{\spacing}

\renewcommand\intextsep{\spacing}

\begin{document}
\title{\textsc{NewsKVQA}: Knowledge-Aware News Video Question Answering}
\author{Pranay Gupta\inst{1} \and
Manish Gupta\inst{1,2}\orcidID{0000-0002-2843-3110}}
\institute{$^1$IIIT-Hyderabad, $^2$Microsoft\\India\\
\email{pranay.gupta@research.iiit.ac.in, gmanish@microsoft.com}}
\maketitle              % typeset the header of the contribution
\begin{abstract}
%business application
Answering questions in the context of videos can be helpful in video indexing, video retrieval systems, video summarization, learning management systems and surveillance video analysis.
%drawbacks of previous methods
Although there exists a large body of work on visual question answering, work on video question answering  (1) is limited to domains like movies, TV shows, gameplay, or human activity, and (2) is mostly based on common sense reasoning.
%technical problem
In this paper, we explore a new frontier in video question answering: answering knowledge-based questions in the context of \emph{news videos}.
%brief approach
To this end, we curate a new dataset of $\sim$12K news videos spanning across $\sim$156 hours with $\sim$1M multiple-choice question-answer pairs covering $8263$ unique entities. We make the dataset publicly available\footnote{\url{https://tinyurl.com/videoQAData}\label{datafootnote}}. Using this dataset, we propose a novel approach, \textsc{NewsKVQA} (Knowledge-Aware News Video Question Answering) which performs multi-modal inferencing over textual multiple-choice questions, videos, their transcripts and knowledge base, and presents a strong baseline.
%8263=sed 's/]}/]}\n/g' mined_facts_for_0_hop_entities.json|wc -l
%results
%\keywords{First keyword  \and Second keyword \and Another keyword.}
\end{abstract}

\section{Introduction}
%History of VQA. Video QA. Knowledge-aware. 
Visual Question Answering (VQA) aims at answering a text question in the context of an image~\cite{antol2015vqa}. The questions can be of various types: multiple choice, binary, fill in the blanks, counting-based~\cite{acharya2019tallyqa}, or open-ended. 
%Several researchers have proposed multiple datasets~\cite{malinowski2015ask,ren2015image,zhu2016visual7w,antol2015vqa,goyal2017making,johnson2017clevr,krishna2017visual,geman2015visual} for VQA.
%Most methods for VQA use either basic multimodal fusion of language and image embeddings~\cite{ren2015exploring,gao2015you,noh2016image,kembhavi2017you},  attention-based multimodal fusion~\cite{yang2016stacked,fukui2016multimodal,shih2016look,lu2016hierarchical,xiong2016dynamic,jabri2016revisiting} or neural module networks~\cite{andreas2016neural,hu2017learning}. More recently, newer problem settings have been proposed as extensions of the basic VQA framework like Text VQA~\cite{singh2019towards}, Common-Sense Reasoning~\cite{narasimhan2018straight}, Visual Dialog~\cite{das2017visual}, Video QA~\cite{zeng2017leveraging}, knowledge-based VQA~\cite{shah2019kvqa} and knowledge-based VQA for videos~\cite{garcia2020knowit}. Extending on this rich literature, we propose a novel problem setting: entity-based question answering in the context of news videos.
Most methods for VQA use either basic multimodal fusion of language and image embeddings~\cite{kembhavi2017you},  attention-based multimodal fusion~\cite{yang2016stacked} or neural module networks~\cite{hu2017learning}. More recently, newer problem settings have been proposed as extensions of the basic VQA framework like Text VQA~\cite{singh2019towards}, Video QA~\cite{zeng2017leveraging}, knowledge-based VQA~\cite{shah2019kvqa} and knowledge-based VQA for videos~\cite{garcia2020knowit}. Extending on this rich literature, we propose a novel problem setting: entity-based question answering in the context of news videos.

%why is news video QA important? Details of applications.
%drawbacks of previous methods
Recently multiple datasets have been proposed for knowledge-based VQA and video VQA. However, many of these questions do not actually make use of the image and knowledge graph (KG) information in a holistic manner. Thus, questions in such datasets can be often answered by using just the image information, or the associated text or a limited KG of dataset-specific entities. We fill this gap by contributing a dataset where each question can be answered only by an effective combination of video understanding, and knowledge-based reasoning. Further, current domain-specific video VQA work is limited to domains like movies, TV shows, gameplay and human activity. In this paper, we focus on the news domain since news videos are rich in real-world entities and facts, and linking such entities to Wikipedia KG and answering questions based on such associations is of immense practical value in video indexing, video retrieval systems, video summarization, learning management systems and surveillance video analysis.

\begin{figure}
    \centering
    \includegraphics[width=0.8\textwidth]{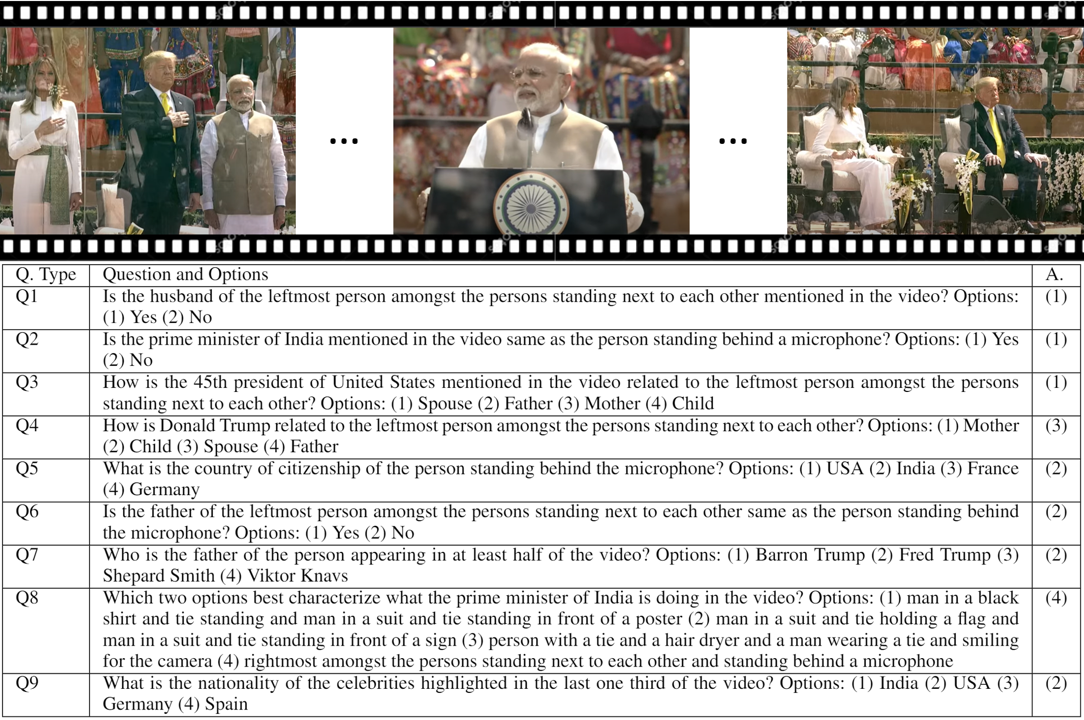}
    \caption{Examples of questions for a video on ``PM Modi and President Trump attends Namaste Trump event in Ahmedabad, Gujarat''\protect\footnotemark. For every question, we show question type (refer Table~\ref{tab:datasetDist}), question, options and answer (A.). }
    \label{fig:posterChildExample}
\end{figure}

%Example
%Could be https://www.youtube.com/watch?v=Noe6hHgq8xY

% dataset contribution
Our dataset gathered from $12094$ news videos spanning 156 hours contains $1041352$ multiple-choice question-answer pairs with links to $8263$ unique Wikipedia entities. All the questions are centered around entities of person type and involve different levels of KG as well as visual reasoning complexity. Questions with high KG complexity need multiple hops of inference over KG nodes. Questions with high visual complexity require multi-frame reasoning, person-person positional relationship inference or person-object relationship inference. We have ensured that incorrect answer options are selected carefully to avoid bias. Fig.~\ref{fig:posterChildExample} shows examples of questions for a video\footref{videofootnote}.
\footnotetext{\url{https://www.youtube.com/watch?v=lEQ2J_Z1DdA}\label{videofootnote}}

% approach, results and insights
For question answering, we present results using a strong baseline which uses a combination of signals across different modes. Given the input video, there are three main sources of knowledge -- text, visual and KG. From the visual part, we can retrieve object boundaries, object names, celebrities, and image representation using popular CNN models like ResNet~\cite{he2016deep}. For the text modality, the question, candidate answer (or option), transcript are used as input. Finally from KG, we can leverage KG facts for entities extracted from transcript (person type entities) or visual frames (using celebrity detection). Text as well as KG signals are processed using pretrained BERT model. Finally, we concatenate image as well as text representations and connect this to a dense layer followed by an output softmax layer to come up with the final prediction. 

%main contributions
In this paper, we make the following main contributions. (1) We build a large dataset for knowledge-aware video QA in the news domain. We make the dataset publicly available\footref{datafootnote}. (2) We carefully design questions and candidate answers such that (a) they need to use both visual information from videos and KG signals to be answered (b) they are diverse with respect to answer types, leverage diverse set of entities, and involve different levels of KG as well as visual reasoning complexity. (3) We experiment with a strong baseline method which leverages different combinations of signals across visual, text and KG modes to achieve good accuracy across various question types.

\section{Related Work}

\begin{table*}[ht!]
%\setlength{\tabcolsep}{5pt}
%\tiny
\centering
\scriptsize
\begin{tabular}{|l|l|c|c|c|l|c|}
\hline
\textbf{Dataset} & \textbf{Domain} & \textbf{\# Video Clips} & \textbf{\# QAs} & \textbf{Answer Type} & \textbf{KG} & \textbf{Entity-Set}\\
\hline
\hline
Youtube2TextQA~\cite{guadarrama2013youtube2text} & Human activity & 1987 & 122708 & MCQ & \xmark & none\\
 \hline
MovieQA~\cite{tapaswi2016movieqa} & Movies & 6771 & 14944 & MCQ & \xmark & closed\\
 \hline
PororoQA~\cite{kim2017deepstory} & Cartoon & 16066 & 8913 & MCQ & \xmark & closed \\
 \hline
MarioQA~\cite{mun2017marioqa} & Gameplay &- & 187757 & MCQ & \xmark & none\\
 \hline
TGIF-QA~\cite{jang2017tgif} & Animated GIFs & 71741 & 165165 & Open+MCQ & \xmark & none\\
 \hline
Movie-FIB~\cite{maharaj2017dataset} & Movies & 118507 & 348998 & Open & \xmark & closed \\
 \hline
VideoQA~\cite{zeng2017leveraging} & Generic & 18100 & 175076 & Open & \xmark & none\\
 \hline
MSRVTT-QA~\cite{xu2017video} & Generic & 10000 & 243680 & Open & \xmark & none\\
 \hline
MSVD-QA~\cite{xu2017video} & Generic & 1970 & 50505 & Open & \xmark & none\\
 \hline
TVQA~\cite{lei2018tvqa} & TV shows & 21793 & 152545 & MCQ & \xmark & closed\\
 \hline
ActivityNetQA~\cite{yu2019activitynet} & Human activity & 5800 & 58000 & Open & \xmark & none\\
 \hline
Social-IQ~\cite{zadeh2019social} & Human Intent & 1250 & 7500 & MCQ & \xmark & none\\
 \hline
KnowIT VQA~\cite{garcia2020knowit} & TV shows & 12087 & 24282 & MCQ & \cmark & closed\\
 \hline
TVQA+~\cite{lei2020tvqa+}&TV shows & 4198 & 29383 & MCQ & \xmark & closed\\
\hline
\hline
\textbf{\textsc{NewsKVQA} (ours)} & News & 12094 & 1041352 & MCQ & \cmark & open \\
\hline
\end{tabular}
\caption{Description of Video Question Answering Datasets}
    \label{tab:videoQADatasets}
\end{table*}

\noindent\textbf{Knowledge-Based Visual Question Answering}: Besides typical multi-modal fusion-based methods, VQA can also be effectively done by combining image semantic representation with external information extracted from a general knowledge base. Initial systems focused on common sense-enabled VQA~\cite{aditya2018explicit,kembhavi2017you,zellers2019recognition}. In this line of work, datasets like KB-VQA~\cite{wang2017explicit}  and FVQA~\cite{wang2017fvqa} were centered around common nouns, and contain only a few images. They are far more restricted in terms of logical reasoning complexity required as compared to later datasets in this area. More recent systems leverage factual information about entities by fusing multi-modal image information with knowledge graphs (KG). Zhu et al.~\cite{zhu2015building} create a specific knowledge base with image-focused data for answering questions under a certain template, whereas more generic approaches like~\cite{wu2016ask} extract information from external knowledge bases, such as DBpedia, for improving VQA accuracy. R-VQA~\cite{lu2018r} has questions requiring information about relational facts to answer. KVQA ~\cite{shah2019kvqa} is a dataset where  multi-entity, multi-relation, and multi-hop reasoning over KGs is needed for VQA. OK-VQA~\cite{marino2019ok} has free-form questions without knowledge annotations. Knowledge-based approaches have mainly focused on fusing KG facts with image representations. We extend this line of work to leverage knowledge from KG facts for answering questions based on \emph{videos}.

\noindent\textbf{Video Question Answering}: Beyond images, recently multiple video question answering datasets have been proposed as described briefly in Table~\ref{tab:videoQADatasets}. To the best of our knowledge, we are the first to propose a video question answering dataset focused on the news domain. Most of the datasets shown in Table~\ref{tab:videoQADatasets} do not use external knowledge from KGs. Only KnowITVQA utilizes in-domain facts about entities in popular sitcoms. Unlike these datasets, we link videos with actual entities in Wikipedia and our proposed dataset contains questions which leverage multi-hop inference on the Wikipedia KG around such entities. Thus, only our dataset has questions related to an open entity set unlike other methods where the set of entities is limited to say characters in sitcoms or none. Also, like many other methods, questions in our dataset are of MCQ (multiple choice questions) form.

%\subsection{Video Analytics}

\section{\textsc{NewsKVQA} Dataset Curation}
\label{sec:datasetintro}
We introduce the novel \textsc{NewsKVQA} dataset for exploring entity based visual QA in news videos using KG. Answering each question requires inferences over multiple modalities including video, transcripts and facts from the KG. The questions exploit the presence of multiple entities in news videos and the spatial-temporal semantics between the objects and the concerned person entities. Questions are further augmented using multi-hop facts about the entities from the KG and the transcripts.

\subsection{Initial Data Collection}
\noindent\textbf{Video Collection}
\label{subsec:vc}
A total of 2730 English news videos are sourced from six Youtube channels of three popular news media outlets: USA today, Al Jazeera and Washington Post such that each video is less than 10 minutes in length. The videos belong to the following domains: entertainment, sports, politics and life. The titles, descriptions and the transcripts are downloaded along with the videos. Each video spans 3.43 minutes on average. These raw videos are converted into $\sim$30 sec clips by combining the consecutive shots detected via shot boundary detection\footnote{ffmpeg implementation of shot boundary detection}. Shots smaller than 30 sec were combined with next shots until a clip of length $\geq$30 sec is obtained. The aligned transcripts are also split according to the shot boundaries. Overall, the dataset contains 12094 clips. Each clip has a transcript with $\sim$78 words on average. Code + data are publicly available\footref{datafootnote}.

\noindent\textbf{Entity Recognition}
\label{subsec:er}
Assuming that the main person entities in video clips would be named in the transcripts, we employ a text entity linker\footnote{The entity linker offered by Microsoft Azure Cognitive Text Analytics Services API} to retrieve the entities from transcripts. This led to $\sim$4.43 entities per clip on average. We retained only those entities with confidence score$>$0.1. Top five entity types were human, business, film, city and country. There were a total of 8263 unique entities of which 1797 are of human type. %Note that there could be entities in the video (called visual entities) which are not mentioned in the transcript. Similarly there could be transcript entities which are just referred in the audio but do not visually appear in the video.

To temporally localize the entities within video clips, we fine-tune a face detection and recognition module\footnote{\url{https://github.com/Giphy/celeb-detection-oss}}, for all the entities retrieved from transcripts. We finetune this model using a collection of 100 images per entity obtained from Google image search. Next, keyframes are sampled with a stride of 3 seconds from the video clips. The bounding boxes and the faces predicted by the face detection and recognition module in each keyframe are stored. In multi-person images, the bounding boxes and the detected entities are sorted in the left-to-right order of their position. Overall, $\sim$62.5\% frames had a person recognized. Of these, $\sim$39.4\% frames had more than one person recognized. 
%Total 154.4 hours means 154.4*60*60/3 frames. Of those, 115805 had a person in celebDetectionData.tsv.

\noindent\textbf{Knowledge Graph Curation}
\label{subsec:kb}
For each of the $8263$ recognized entities, we obtain facts as well as linkage information from Wikidata\footnote{\url{https://www.wikidata.org/wiki/Special:EntityData/<entityQID>.json}}. Further, we do a multi-hop traversal on the Wikidata KG to materialize up to two hop facts related to each of our entities. We store the facts as \{subject entity, relation, object entity\} triplets $\langle e_1,r,e_2\rangle$, wherein the entities are marked by unique Wikidata item IDs (QIDs) and the relations are marked by unique property IDs (PIDs). Overall our KG has $42848$ entities and $430985$ relations. 
%sed 's/]}/]}\n/g' mined_facts_for_0_hop_entities.json |wc -l
%8263
%sed 's/]}/]}\n/g' mined_facts_for_1_hop_entities.json |wc -l
%34585
%34585+8263
%42848
%sed 's/]}/]}\n/g' mined_facts_for_0_hop_entities.json |awk -F"\"facts\":" '{print $2}'|sed 's/\], \[/\n/g'|grep -v "\[\]"|wc -l
% 88966
% sed 's/]}/]}\n/g' mined_facts_for_1_hop_entities.json |awk -F"\"facts\":" '{print $2}'|sed 's/\], \[/\n/g'|grep -v "\[\]"|wc -l
% 342019
% 88966+342019
% 430985
\subsection{Question Generation}
\label{subsec:qg}
\begin{figure}
\begin{minipage}{0.48\textwidth}
\centering
\scriptsize
\begin{tabular}{|l|l|l|l|c|}
\hline
Question&Answer&Source of&Answer&\#Questions\\
Type&Evidence&Q. entities&Type&\\
\hline
\hline
Q1& V+T+KG & V &Binary&29291\\
\hline
Q2& V+T+KG & V+T &Binary&95506\\
\hline
Q3& V+T+KG & V+T &Relation&23245\\
\hline
Q4& V+KG & V+KG &Relation&21207\\
\hline
Q5& V+KG & V &Entity&598762\\
\hline
Q6 & V+KG & V &Binary&52156\\
\hline
Q7 & V+KG & V &Entity&59536\\
\hline
Q8 & V+KG & V &Desc.&48160\\
\hline
Q9 & V+KG & V &Entity&113489\\
\hline
    \end{tabular}
    \captionof{table}{Characterization of Various Types of Questions in the \textsc{NewsKVQA} Dataset. V=Visual, T=Transcript, KG=Knowledge graph.}
    \label{tab:datasetDist}
\end{minipage}
\hspace{0.02\textwidth}
\begin{minipage}{0.48\textwidth}
    \centering
    \includegraphics[width=0.6\textwidth]{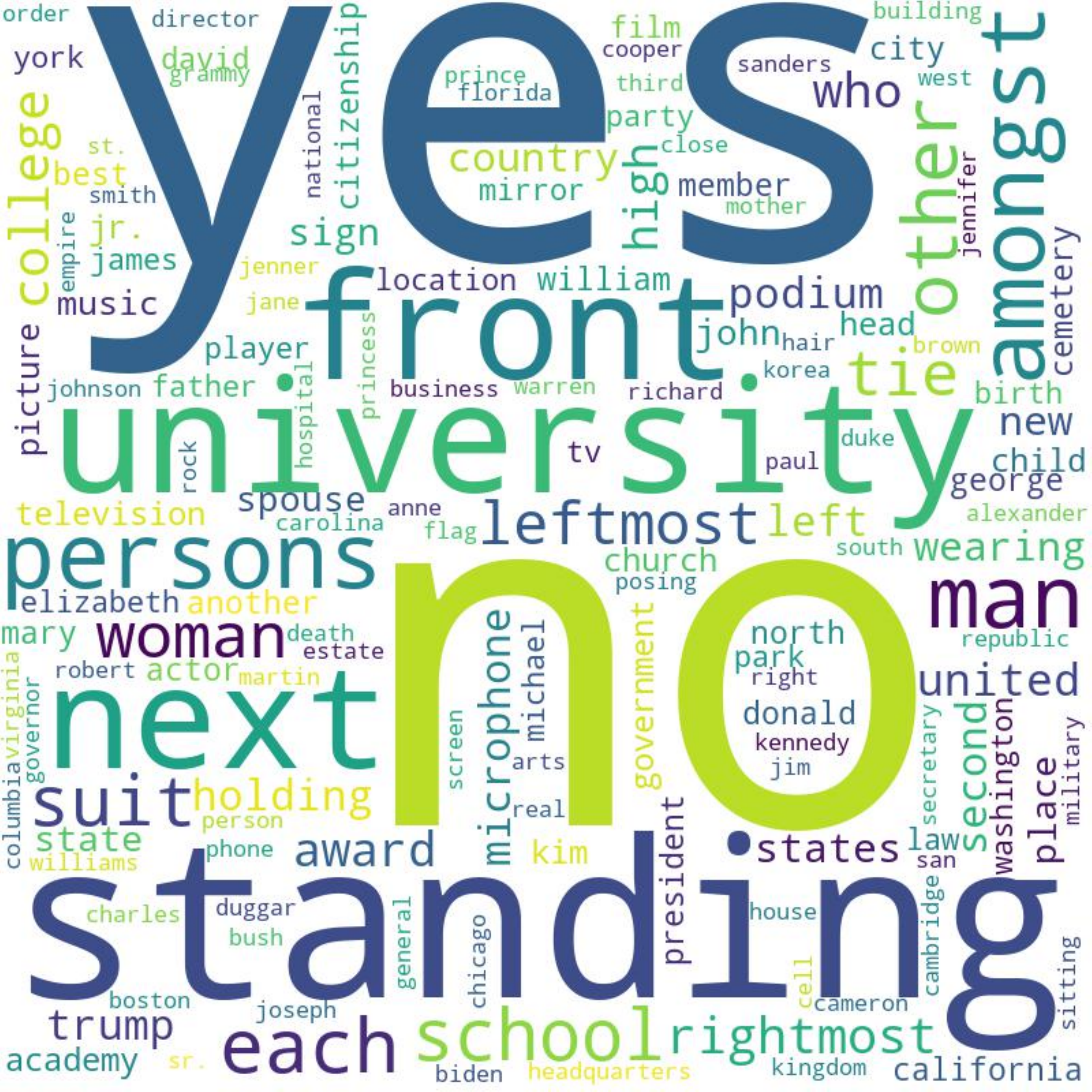}
    \caption{Word Cloud of answer words and phrases in our \textsc{NewsKVQA} dataset}
    \label{fig:wordCloud}
\end{minipage}
\end{figure}

Manual question generation would be slow and difficult to scale. Owing to the development of reliable image captioning systems~\cite{li2020oscar}, we were able to substantially bring down the human effort in generating the questions corresponding to the video clips. Automating this procedure helped us efficiently generate questions at scale. Table~\ref{tab:datasetDist} shows a  basic characterization of the 9 question types in our \textsc{NewsKVQA} dataset. All the question types require inferences over data from multiple modalities like visual, textual and KG facts. Some questions need visual cues in terms of person-person while others need to understand person-object relationships from videos. %E.g., in the group Q1, Q1a captures person-object relationship where only one person is involved (SPO=Single Person Object), Q1b captures person-object relationship where multiple persons are involved (MPO=Multi Person Object) while Q1c captures person-person relationship (MPP=Multi Person Person).
Answer types could be entities, relations, binary, or person description. In addition, answering complexity varies in terms of number of KG hops across different question types. In this section, we discuss our detailed methods for generating different types of questions.

%In this section, we discuss caption generation and template creation, the process for generating questions needing video+KG, the process for generating questions needing video+transcript+KG, and option generation.

% \begin{table}
% \centering
% \scriptsize
% \begin{tabular}{|l|l|l|l|c|}
% \hline
% Question&Answer&Source of&Answer&\#Questions\\
% Type&Evidence&Q. entities&Type&\\
% \hline
% \hline
% Q1& V+T+KG & V &Binary&29291\\
% \hline
% Q2& V+T+KG & V+T &Binary&95506\\
% \hline
% Q3& V+T+KG & V+T &Relation&23245\\
% \hline
% Q4& V+KG & V+KG &Relation&21207\\
% \hline
% Q5& V+KG & V &Entity&598762\\
% \hline
% Q6 & V+KG & V &Binary&52156\\
% \hline
% Q7 & V+KG & V &Entity&59536\\
% \hline
% Q8 & V+KG & V &Desc.&48160\\
% \hline
% Q9 & V+KG & V &Entity&113489\\
% \hline
%     \end{tabular}
%     \captionof{table}{Characterization of Various Types of Questions in the \textsc{NewsKVQA} Dataset. V=Visual, T=Transcript, KG=Knowledge graph. Detailed table in Appendix.}
%     \label{tab:datasetDist}
% \end{table}

\noindent\textbf{Generating questions with entities as answer type}
% single and multiperson, single frame object-person relationship, time augmentation
For each video keyframe, we obtain the respective captions via Oscar caption generator~\cite{li2020oscar}. The caption captures the visual relation of the person with an object or other persons in the image. E.g., `A person standing behind the microphone', `A group of people standing next to each other'. Thus, for every entity, there are 2 descriptions: surface form from transcript (which we call transcript entity) and visual description from captions (which we call visual entity).

%Exploiting the common structure of the obtained captions, we identify the subject of the sentence. 
We assume that the subject (grammatical) in the caption is the entity recognized by the celebrity detector in the keyframe. We modify the subject (grammatical) within the caption with a templatized question about the relation in the knowledge base associated with the entity. E.g., a  generated caption, `\textit{A person} standing behind a microphone', is changed to, `\textit{Who is the mother of the person} standing behind the microphone'. Question templates for question related to every type of relation (relevant to a person entity) are manually created. In total, we have 30 such templates. Further if multiple keyframes within a video have the same caption, the question is augmented using the time of appearance of the keyframe. E.g., `Who is the mother of the person standing behind the microphone \textit{in the beginning one third of the video?}'. For multi-person keyframes, we leverage the knowledge of the spatial ordering of the entities in the keyframe  and use positional indicators like `leftmost', `second to the rightmost', etc. to refer to the person being questioned about in the keyframe. E.g., `Who is the mother of the \textit{second from the left} amongst the persons standing next to each other in the beginning one third of the video?' All these heuristics help us obtain questions of type Q5 where the question is centered around visual entity (i.e., entity described in the video) using (1) hints from the generated captions, and (2) facts from KG. %The three types differ in the relationship captured between person(s) and objects (similar to the way Q1 differ) in keyframes. 
Formally, given a KG fact, $\langle e_1,r,e_2\rangle$, Q5 questions were formed using visual entity (whose description is obtained from captions) as $e_1$, KG based relation as $r$ and expected $e_2$ as the answer. 

\noindent\textbf{Generating questions with binary answer type}
Given a KG fact, $\langle e_1,r,e_2\rangle$, another way to frame questions is to check if $e_2$ is mentioned in the video or transcript, given visual entity $e_1$ and $r$ in the question. This leads to questions of type Q1. 
E.g, `Is the mother of the person standing behind the microphone mentioned in the video?' 
Further, more complex questions (such as of type Q6) can be designed such that given two facts $\langle e_1',r',e_2'\rangle$ and $\langle e_1'',r'',e_2''\rangle$, visual entities $e_1'$ and $e_1''$ appearing in two different keyframes in the same video, the question may contain $e_1'$, $e_1''$, $r'$ and $r''$ and asks to check if $e_2'$ and $e_2''$ are the same. E.g, `Is the occupation of the rightmost amongst the persons standing in a room with a clock on the wall the same as the occupation of the rightmost amongst the persons standing next to each other?'

\noindent\textbf{Generating questions with relation as answer type}
% relation as answer
%The facts in the knowledge base are stored as \textit{(entity1, relation, entity2)} triplets. Until now we formed the questions, using each of the detected entities in the keyframe as entity1, asking about entity2 based on the relation. 
Questions that expect $r$ as answer can be created by including both $e_1$ and $e_2$ in the question. While we fix $e_2$ to be always visual entity (to ensure that all our questions need video for answering), $e_1$ can be a KG entity like in questions of type Q4. E.g., `\textit{How is Paris related to the country of citizenship of} the person standing behind the mic?'.
%Questions that expected $r$ as answer can be created by including both $e_1$ and $e_2$ in the question. While, we fix $e_2$ to be always video entity (to ensure that all our questions need video for answering), $e_1$ can be a KG entity or a transcript entity. In questions of type Q10-Q12, the $e_1$ is directly taken from KG while $e_2$ corresponds to the video entity. In questions of type Q7-Q9, the $e_1$ refers to an attribute (obtained from KG) of an entity extracted from transcripts while $e_2$ corresponds to the video entity. Often times, transcripts entities are not visualized in the entity, and hence are very different compared to video entities.

%This can be reversed to generate more questions, i.e. asking about the relation based on the entity2. 
%For ex. `\textit{How is Washington DC related to the country of citizenship of} the person standing behind the microphone?`. However, 
The choice of $r$ for generating such questions needs to be done carefully to avoid trivial/obvious questions. 
%Propagating this strategy to all the relations within all the entities result in questions with obvious answers. 
Consider these examples: (1) `What is the relation between North America and the country of the citizenship of the person behind the microphone?' with the answer `\textit{Continent}'. (2)  `What is the relationship between Fordham University and the person standing behind the microphone?' with answer `\textit{educated at}'. In the KG, `North America' is only related to other entities as a `continent', similarly `Fordham University' is related to other person entities only as an `alma mater'. Hence, we prune away questions where $(t_1, t_2)$ is very frequently related to relation $r$ (rather than other relations) where $t_1$ and $t_2$ are types of entities $e_1$ and $e_2$ resp.

%\subsubsection{Generating questions needing video + transcript + KG}
\noindent\textbf{Generating questions involving transcript entities}
%In the previous part, we discussed questions Q7-Q9 which contain a transcript entity as part of the question, and expect a relation answer type. Further, questions Q4-Q6 also contain a video entity and a transcript entity in the question but answer type is binary.  
Sometimes, transcripts contain (non-person) entities which are not visualized in the video, and hence transcripts can mention novel entities. We combine such transcript entities with visual entities to create questions where answer type is binary (Q2) or relation (Q3). 

% subtitle entities mentioned in video
%Using the entities detected in the transcript \ref{subsec:er}, we can create questions about the relationship between the transcript entities and the entities recognized by the celebrity detector in the keyframes.
The first set of questions, inquire whether the visual entities are related to the transcript entities. Thus, $e_1$, $e_2$ and $r$ are mentioned in the question, and the answer could be Yes or No. E.g., `Is the country of citizenship of the person standing behind the microphone mentioned in the video?' We create similar questions using multi-hop relationships of the transcript entities and visual entities. E.g., `Is the father of the $\langle$transcript entity$\rangle$ mentioned in the video, same as the person standing behind the microphone?' However directly using the transcript entity in the question would trivialize the answering process, therefore we use a representative description of the transcript entity. For this, we leverage entity descriptions provided as part of our KG. Thus the question is converted to, `Is the father of the 45th president of the USA same as the leftmost amongst the persons posing for a picture?'. Actor/actress entities often have very similar descriptions in the KG, hence for such entities, the description is augmented with information about their latest movies/shows. E.g., for `Steve Carell', the generated question would be `Is the child of the American actor who acted in the movie The Office, same as the leftmost amongst the persons standing next to each other in the video?' 

%In addition to this, questions inquiring about the relationship between the transcript entities and the subtitle entities can also be created. 
The second set of questions mention transcript entity as $e_1$ and visual entity as $e_2$ in the question and seek relation $r$ as the answer. E.g., `How is the 45th president of the United States mentioned in the video related to the country of citizenship of the leftmost amongst the persons posing for a picture?' with the answer \textit{`Head of Government'}. We randomly remove the questions about the most frequent transcript entities as a bias correction step.

\noindent\textbf{Generating questions based on video durations}
%temporal questions
%The questions created until now utilise visual information from within a keyframe and thus span only a segment of the video. To add to the complexity of the questions, 
We further create 3 types of questions (Q7, Q8, Q9) about entities spanning across larger video durations. 
%We first create binary questions about the entities across different keyframes of a video by merging their captions and checking for similarity in terms of their relations. For ex. ` Is the occupation of the rightmost amongst the persons standing in a room with a clock on the wall the same as the occupation of the rightmost amongst the persons standing next to each other?`. 
% \begin{figure}
%     \centering
%     \includegraphics[width=0.8\columnwidth]{wordCloud}
%     \caption{Word Cloud of answer words and phrases in our \textsc{NewsKVQA} dataset}
%     \label{fig:wordCloud}
% \end{figure}
Firstly, we create questions (of type Q7) about the entities, which appear more prominently in the video, e.g., `What is the alma mater of the celebrity appearing in more than half of the video?'. 

Secondly, we create questions (of type Q8) regarding what each entity in the video is doing across different keyframes in the video. Similar to the transcript entity questions, the keyframe entity is referenced in the question using the description. E.g., `Which statement best characterizes the American internet personality present in the video?' with answer: `\textit{leftmost amongst the persons standing in a room with a clock on the wall and person with yellow shirt holding a mic}'. The answer is extracted from the keyframe captions, expressing the entity relationship with the object and/or other persons in the keyframe. Of course, we omit keyframes with same captions. 

Finally, we create questions (of type Q9) about the most common relation among the entities appearing within a span of the video. E.g., `What is the nationality of the celebrities highlighted in the beginning one third of the video?' The common nationality of more than 50\% of the entities within the span is chosen as the answer. If less than 50\% of the entities have the common entity then the question is removed.

\begin{figure}
    \centering
    \includegraphics[width=0.8\textwidth]{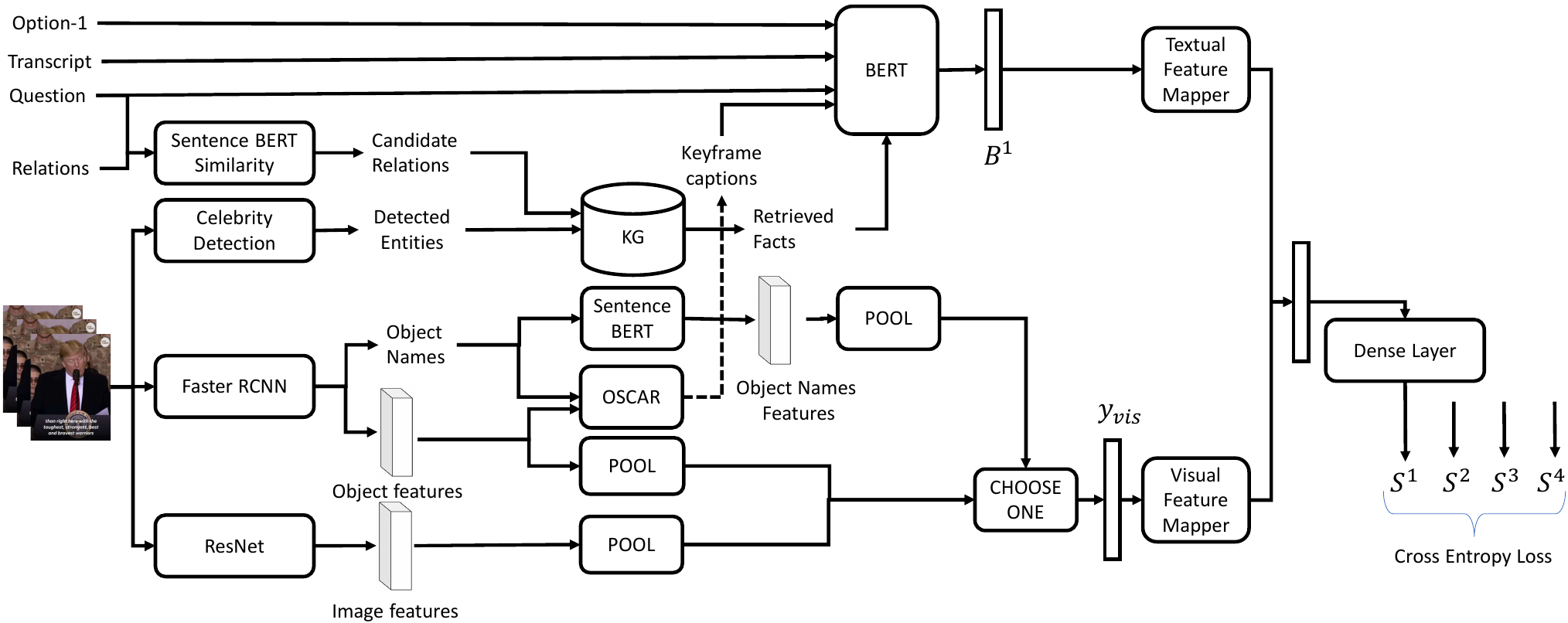}
    \caption{\textsc{NewsKVQA} System Architecture}
    \label{fig:approach}
\end{figure}

\subsection{Option generation}
We select 3 options for each of the generated questions in addition to the correct answer, however the procedure for generating the options differs according to the question type. For Q5, where the answer is an entity, other entities of the same entity type as the correct answer are chosen as options. E.g., for the correct answer ``United States'', other countries are chosen as options. Since, randomly choosing the options adds a bias towards the more frequently occurring correct answers, entities which are approximately as frequent as the correct answer are chosen as options. The max frequency with which an entity can be a correct answer is set to $300$.

For Q4, where answer type is a relation, the other relations associated with the entity whose relation is being asked about in the question are used as options. E.g., for the question ``How is the 45th president of the USA mentioned in the video related to the leftmost person amongst the persons standing in front of the mic?'', the options can be father, child, spouse, head of government, etc. In case less than 4 relations are present for the entity in discussion, then the remaining options are chosen randomly from the set of all possible relations. For binary questions, ``yes'' and ``no'' are the only two options. Similarly for the final person description questions the options are chosen randomly.

\subsection{Dataset Analysis}
\label{sec:datasetdist}

\begin{figure}
%\hspace{-0.05\textwidth}
\begin{minipage}{0.18\textwidth}
   \centering
    \scriptsize
    \begin{tabular}{|l|c|c|}
    \hline
    Type&Yes&No\\
\hline
\hline
Q1&14749&14542\\
\hline
Q2&49351&46155\\
\hline
Q6&26078&26078\\
\hline
    \end{tabular}
    \captionof{table}{Distribution of answers (Yes/No) for questions with binary answers.}
    \label{tab:binAnswerDist}
\end{minipage}
\hspace{0.01\textwidth}
\begin{minipage}{0.24\textwidth}
        \includegraphics[width=\columnwidth]{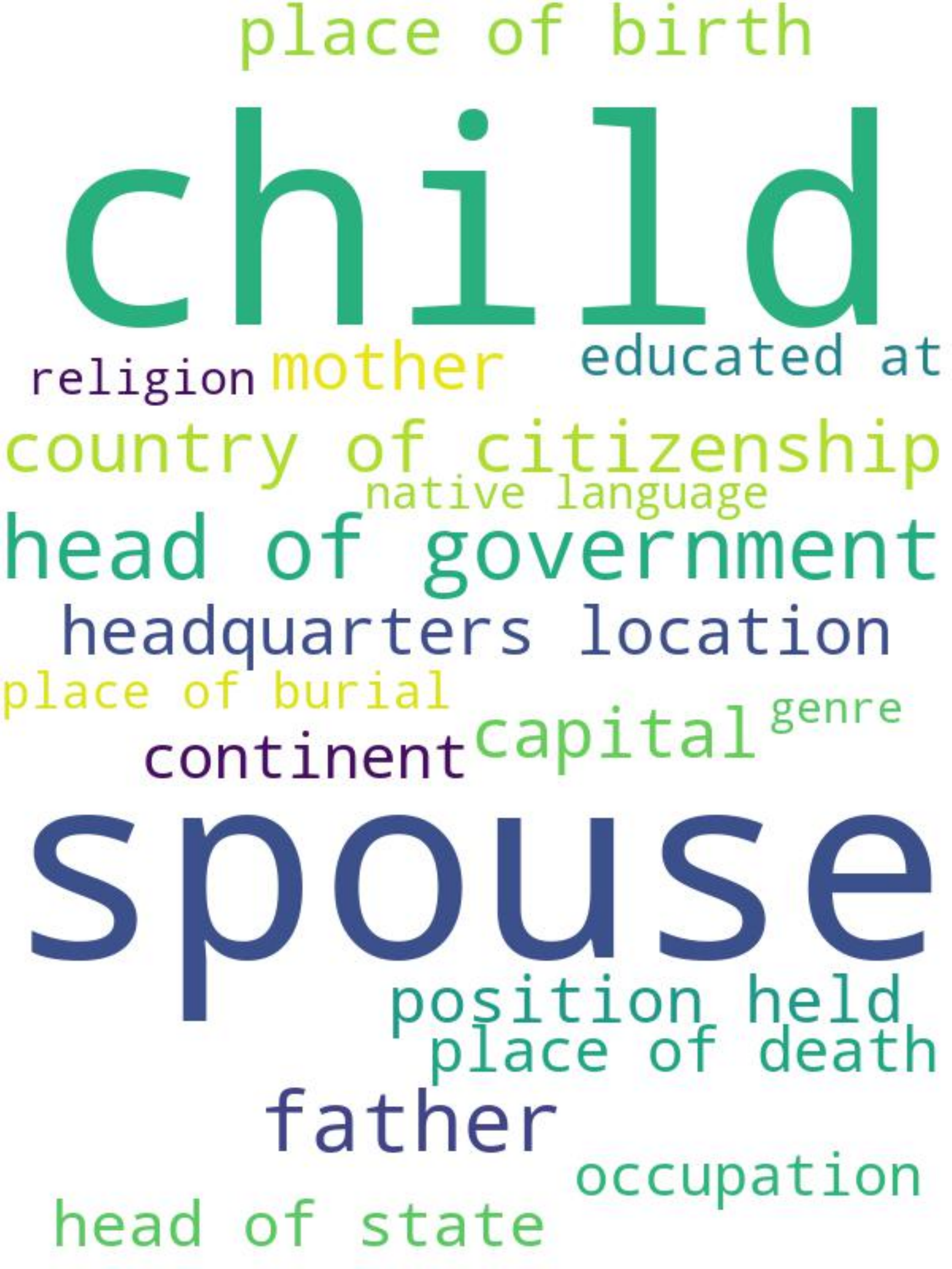}
    \caption{Unique answers for questions of type Q3.}
    \label{fig:wordCloudRel7}
\end{minipage}
\hspace{0.01\textwidth}
\begin{minipage}{0.24\textwidth}
    \includegraphics[width=\columnwidth]{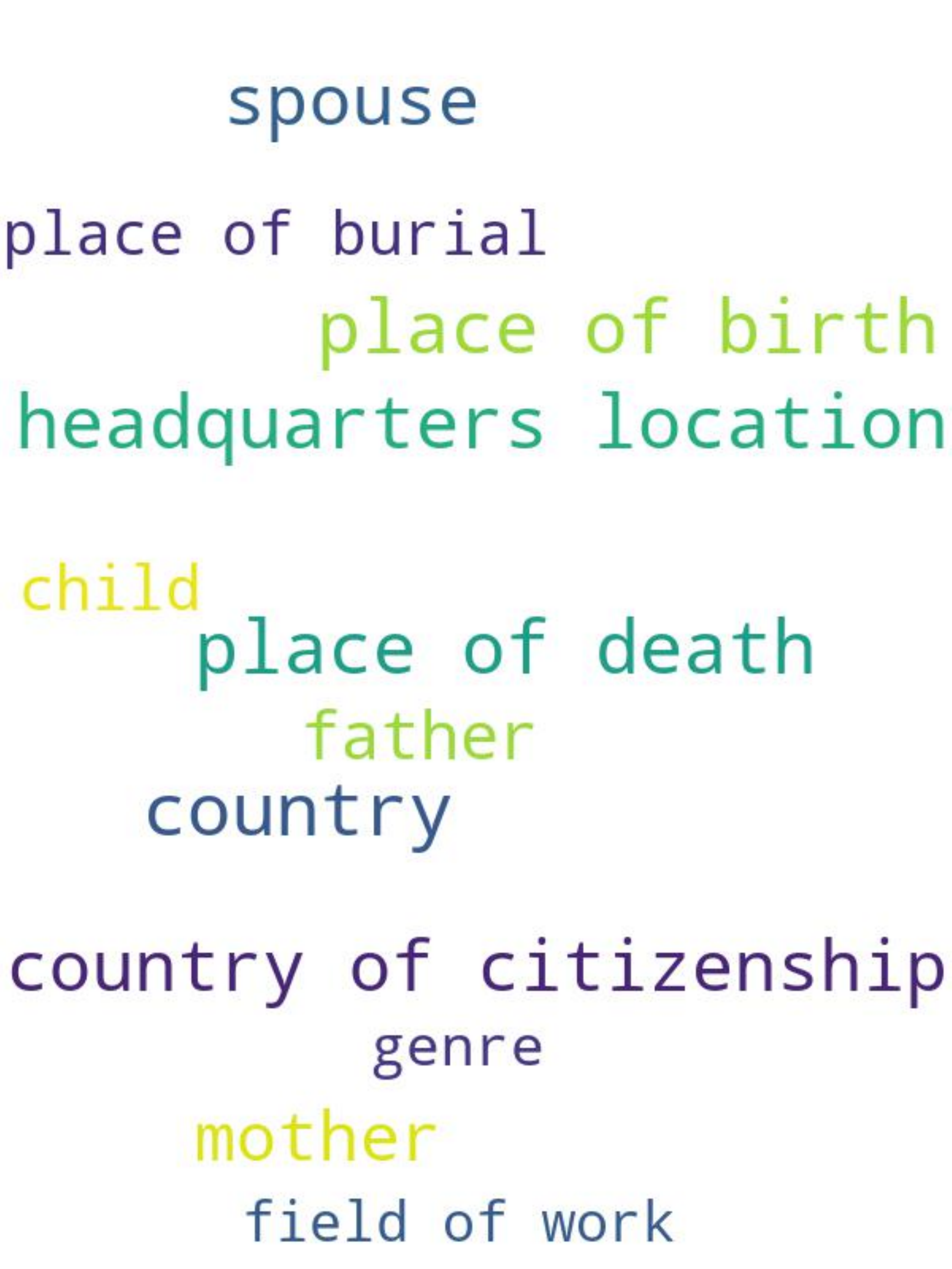}
    \caption{Unique answers for questions of type Q4.}
    \label{fig:wordCloudRel10}
\end{minipage}
\hspace{0.01\textwidth}
\begin{minipage}{0.24\textwidth}
    \includegraphics[width=\columnwidth]{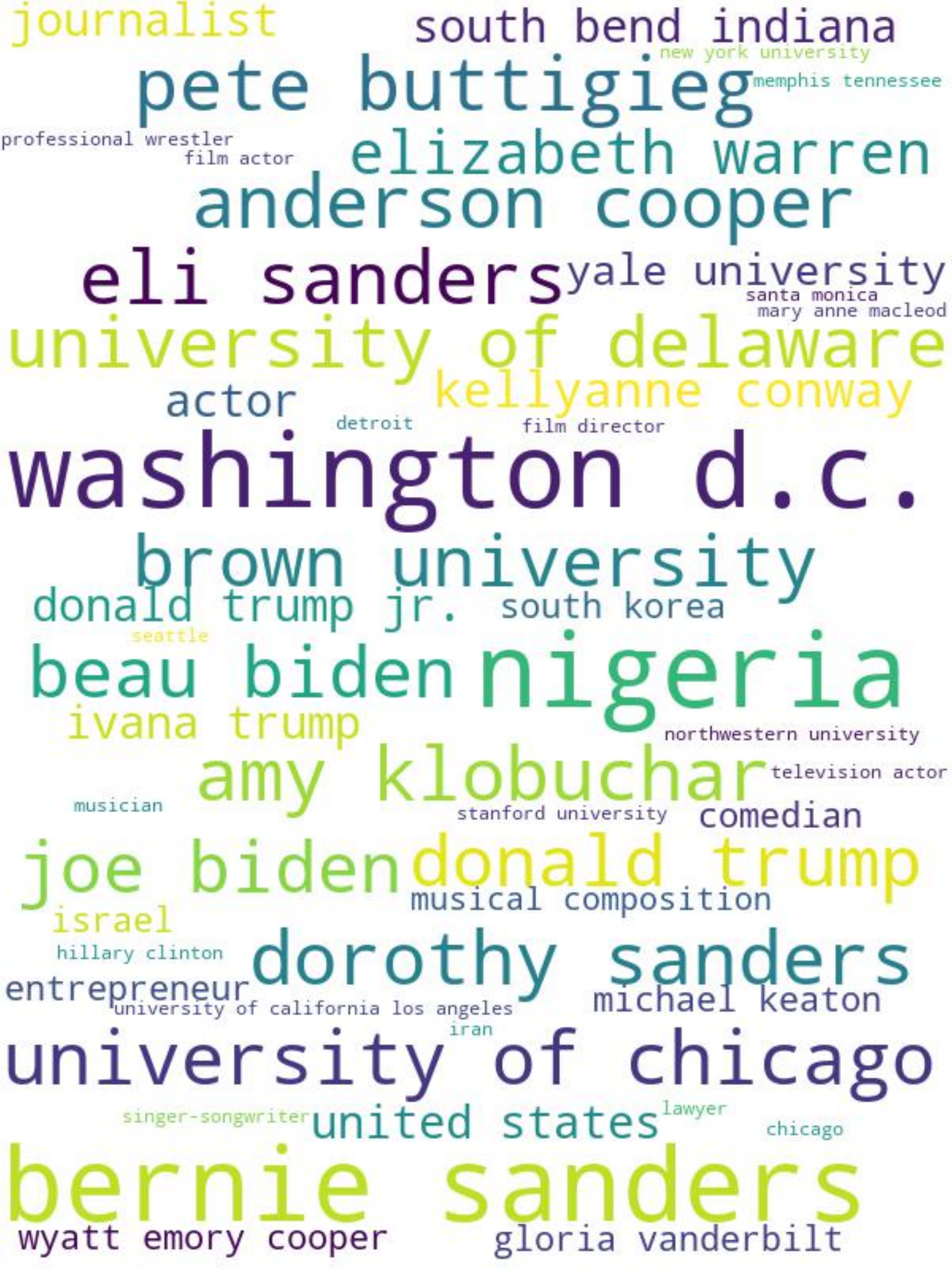}
\caption{150 most frequent unique answers for type Q5.}
    \label{fig:WordCloudEntity1}
\end{minipage}
\end{figure}

On average, each clip has 86.10 questions and each question contains 24.02 words.  
% sed 's/]}/]}\n/g' mined_facts_for_0_hop_entities.json |awk -F"\"facts\":" '{print $2}'|sed 's/\], \[/\n/g'|grep -v "\[\]"|cut -d'"' -f2|sort|uniq > tmp.txt
% sed 's/]}/]}\n/g' mined_facts_for_1_hop_ent
% ities.json |awk -F"\"facts\":" '{print $2}'|sed 's/\], \[/\n/g'|grep -v "\[\]"|cut -d'"' -f2|sort|uniq >> tmp.txt
% sort tmp.txt |uniq |wc -l
% 100
Also, the KG contains information about 4400 different entity types. %types: 4400
%cat mined_facts_for_*|sed 's/"instance of"/\nXYZABC/g'|grep "XYZABC"|cut -d'"' -f2|sort|uniq |wc -l
%4400
Other basic statistics related to our dataset are as follows: transcript size is 84.62 words, answers are on average 3.14 words long, average question size is 22.33 words, average number of facts per entity in our KG is 10.83 (including instance-of relationships).
%sed 's/"facts": /\n/g' mined_facts_for_0_hop_entities.json |cut -d'}' -f1|grep "^\["|sed 's/\], \[/###/g'|awk -F'###' '{print NF}'|awk '{ sum += $1 } END { if (NR > 0) print sum / NR }'
%10.83
Table~\ref{tab:datasetDist} reports the distribution of questions within each category. Fig.~\ref{fig:wordCloud} shows the word cloud of various answer words and phrases across all question types in our dataset. Table~\ref{tab:binAnswerDist} shows the distribution of binary answers across different question types. We show a word cloud of answers for questions with relation (Figs.~\ref{fig:wordCloudRel7} and~\ref{fig:wordCloudRel10}) and entity (Fig.~\ref{fig:WordCloudEntity1}) answer types.
%To perform a quality check for the generated questions, we manually verified 200 questions and found $\sim$10\% of them to be erroneous or ill-formed because of errors either in captions or in celebrity identification.

\section{Knowledge-Aware Video QA Methodology}
\label{sec:method}

We first delineate our approach for question answering on our proposed \textsc{NewsKVQA} dataset. As shown in Figure~\ref{fig:approach}, we first extract relevant facts from the KG using the question and the entities recognized in the video. The answering module then processes these facts along with the question, transcripts and the generated captions.  

%Cite related papers: BERT~\cite{devlin2019bert}, Faster-RCNN~\cite{ren2015faster}, Image captioning~\cite{li2020oscar}, celebrity detection.

\noindent\textbf{Knowledge Base Fact Retrieval}
Given a question and its corresponding video, relevant facts can be retrieved using the entities present in the video and the property around which the question has been framed. Thus we begin by identifying celebrities present in the video. %Similar to the dataset curation step\ref{subsec:er}, we use our finetuned person identifier, to identify entities within all the keyframes of the video. Keyframes, are chosen from the video with a stride of 3 seconds (3*fps frames). 
For extracting relevant properties, we rank all the available properties based on their similarity scores with the question. The properties with the maximum similarity scores are chosen as candidate properties. The similarity scores are calculated as the cosine distance between the Sentence-BERT~\cite{reimers2019sentence} embedding of the question and the properties. Both one and two hop facts pertaining to the detected entities and the candidate properties are extracted from the KG.

\begin{table*}
    \centering
    \scriptsize
    \begin{tabular}{|p{0.9in}|l|c|c|c|c|c|c|c|c|c|}
    \hline  
Info. used&Method&Q1&Q2&Q3&Q4&Q5&Q6&Q7&Q8&Q9\\
\hline
\hline
&Random & 50 & 50 & 25 & 25 & 25 & 50 & 25 & 25 & 25\\
\hline
\multirow{2}{*}{Options}&Longest &50.63&50.03&32.99&25.55&24.53&50.15&26.5&27.41& 25.54\\
\cline{2-11}
&Shortest &50.65&50.03&24.82&23.39&24.92&50.15&25.68&21.06& 25.34\\
\hline
\multirow{2}{*}{Question + option}
%& word2vecSim &&& & & && & &\\
%\cline{2-11}
&BERTSim & 57.97 & 51.84 & 48.05 & 40.11 & 28.22 & 51.94 & 29.94 & 27.31 & 30.45\\
\cline{2-11}
&QA &50.35&50.53&24.45&25.84&24.37&50.24&23.65&26.87& 23.97 \\
\hline
\multirow{2}{1in}{Transcript + question + option}&BERTSim QTA& 53.17 & 55.12 & 38.57 & 34.99 & 28.19 & 54.21 & 29.89 & 26.62 & 29.96 \\
\cline{2-11}
&QTA &50.2&50.05&28.66&24.44&25.27&50.24&25.36&26.68& 25.89 \\
\hline
\multirow{4}{1in}{Visual + transcript + question + option}&QTCA &50.46&50.66&26.93&25.5&24.94&50.06&25.36&25.73& 25.62 \\
\cline{2-11}
&\textsc{QTOnA} & \textbf{57.34} &50.51&74.99&57.84&33.23&53.09&37&39.66& 34.80\\
\cline{2-11}
&\textsc{QTOfA} &56.81&53.33&33.54&23.68&25.02&56.31&25.29&25.67& 26.15\\
\cline{2-11}
&QTRA &54.65&52.12&34.84&22.58&24.78&56.85&24.51&24.55& 25.16\\
\hline
\multirow{3}{1in}{KG facts + visual + transcript + question + option} & QTCFA &53.28& \textbf{56.42} & \textbf{84.41} & \textbf{67.60} &42.71& \textbf{62.04} & 51.85 & 47.91 & 45.18 \\
\cline{2-11}
%&\textsc{QTCOnFA} &50.59&50.53&51.08&39.94& 51.61 &50.22&51.04&76.62& 66.92\\
%\cline{2-11}
&\textsc{QTCOnFA} &50.59&51.98&64.76&52.57& \textbf{54.27} &51.41&\textbf{55.39}&\textbf{81.32}& \textbf{68.28}\\
\cline{2-11}
%&\textsc{QTCOfFA} &50.5&55.37&55.62&38.3&51.36&52.04&50.42&78.82& 66.67\\
%\cline{2-11}
&\textsc{QTCOfFA} &51.66&54.36&64.09&51.69&52.39&51.66&53.09&80.78& 67.28\\
\hline
% ViLBERT&&&&&&&&&&\\
% \hline
% MAG&&&&&&&&&&\\
% \hline
% ViLBERT + KG facts&&&&&&&&&&\\
% \hline
% MAG + KG facts&&&&&&&&&&\\
% \hline
    \end{tabular}
    \caption{Main Results. Q=Question, T=Transcript, C=Caption, A=Answer, \textsc{Of}=Object features, \textsc{On}=Object names, F=facts, R=ResNet.}
    \label{tab:mainResults}
\end{table*}

\noindent\textbf{Visual Information Extraction}
%%Since, the questions in the dataset refer to the facts in the knowledge base along with visual cues, it is imperative that we use visual information for question answering. We extract visual information in multiple ways. 

\noindent{(1) Image Features}: For each of the keyframes, we extract the image features using an ImageNet-pretrained ResNet model. The features are average-pooled along the temporal dimension to obtain the final visual embedding ($y_{vis}$).   

\noindent{(2) Object Names}: A pretrained Faster-RCNN~\cite{ren2015faster} outputs objects within each keyframe. For each frame, we obtain $10$ objects detected with the highest confidence scores. Then we select $10$ most frequent objects from these candidates. For each of these object we retrieve their Sentence-BERT~\cite{reimers2019sentence} embeddings. 

\noindent{(3) Object Features}: We use the final layers features of Faster-RCNN for the $10$ most frequent and confidently detected objects. To add positional information about the detected objects, we concatenate the features with the normalized bounding box coordinates of the detected object. For both object names and object features, the features are first average-pooled along the temporal dimension and then concatenated along the object dimension to obtain the final embedding. Note that image features, object names or object features can be used to calculate $y_{vis}$.

\noindent{(4) Image Captions}: We generate captions of the keyframes using Oscar~\cite{li2020oscar}. 

\noindent\textbf{Answering Module}
Let $q$ denote the input question, $t$ be the transcript, $c$ be the concatenated captions for all keyframes, $f$ be all the concatenated KG facts, $o^{i}$ be the $i^{th}$ option. For $i \in (1, 2, 3, 4)$, we obtain BERT embeddings ($B^{i}$), using $y_{text}^i=[CLS] + q + t + c + [SEP] + f + o^{i} + [SEP]$ as input. We pass $B^{i}$ through the text mapper and $y_{vis}$ through the visual-information mapper. Both the mappers are one dense layer each. The text and visual-information mapper outputs are concatenated before being projected to a single score $s^{i}$. The score for the correct option out of the 4 ($s^{1}$, $s^{2}$, $s^{3}$, $s^{4}$) are maximized under cross entropy loss. 

%\textcolor{red}{We may need to group results by type, or 1-hop vs 2-hop, etc. to see what looks reasonable.}

% \begin{table}
% \setlength{\tabcolsep}{5pt}
% \renewcommand{\arraystretch}{1.3}
% \tiny
% \resizebox{\columnwidth}{!}
%  {
%   \centering
%  \begin{tabular}{lcccc}
%  \toprule
%              Method & 1 hop smp & 2 hop smp &  av  & all\\
%  \midrule
%   bert QA & $xx$ & $xx$ & $xx$ & $xx$\\
%   Our approach QA & $34.4$ & $35.46$ & $50.05$ & $xx$\\
%  \midrule
%   Our approach QOnA & $43.41$ & $xx$ & $xx$ & $xx$\\
%   Our approach QOfA & $43.94$ & $xx$ & $xx$ & $xx$\\
%   Our approach QCA & $xx$ & $xx$ & $xx$ & $xx$\\
%   \midrule
%   Our approach QCFA & $61.44$ & $47.09$ & $53.40$ & $xx$\\
%   Our approach QCFOnA & $xx$ & $xx$ & $xx$ & $xx$\\
%   Our approach QCFOfA & $xx$ & $xx$ & $xx$ & $xx$\\
  
%   \bottomrule
%  \end{tabular}
%   }
% \caption{ Increasing difficulty level of the dataset.}
% \label{tab:news-vqa-baselines}
% \end{table}

\noindent\textbf{Reproducibility Details} We use a pretrained BERT-large model to obtain a $1024$D embedding $B^{i}$. For $y_{vis}$, with image features we get a $2048$D embedding. We use 10 objects per keyframe for objects based embeddings and obtain $10240$D (1024$\times$10) embedding for Object names and $20560$D ((2048 (faster-RCNN output) + 8 (positional embedding)) $\times$10 (\#objects)) embedding for object features. Both textual mapper and visual-information mapper are dense layers that map $B^{i}$ and $y_{vis}$ to $512$D vectors. All linear layers outputs are activated by ReLU. The score $s_{i}$ is passed through a sigmoid activation. We finetune the final 2 layers of the BERT model along with the dense layers using Adam optimiser with a learning rate of $0.0001$. The experiments were run with a batch size of 175 on a machine with 8 V100 GPUs with 32 GB RAM. Each experiment was run for 5 epochs. Note that we also make our code and dataset publicly available\footref{datafootnote}.

\section{Experiments and Results}
\label{experiments}
We perform extensive experimentation on our \textsc{NewsKVQA} dataset using the following methods. %We test with various combinations of input data modalities to understand their importance in answering questions in our dataset.
(1) \textbf{Options}: We simply choose the longest option (Longest) or the shortest option (Shortest) as the answer.
(2) \textbf{Question+Option}: In BERTSim, options with max cosine similarity with the question are selected. The embeddings are calculated using Sentence-Bert. QA is a simplification of our approach (Fig.~\ref{fig:approach}) wherein only the question and the options are used to obtain $y_{text}^{i}=[CLS] + q + [SEP] + o^{i} + [SEP]$. BERT embedding $B^{i}$ is directly projected to get the score $s^{i}$.
(3) \textbf{Transcript+Question+Option}: In addition to the questions and the options the transcripts are also used to identify the correct answer. In BERTSim QTA, option with the maximum cosine similarity with the question and the transcripts is selected as the answer. QTA is an extension of QA wherein the question, the transcripts and the options are used to obtain $y_{text}^{i}= [CLS] + q + t + [SEP] + o^{i} + [SEP]$.
(4) \textbf{Visual+Transcript+Question+Option}: We use visual information in addition to the question, option and transcript to determine the correct answer. In QTCA, visual information is conveyed using  keyframe captions. Captions are fed through BERT as $y_{text}^{i}=[CLS] + q + t + c + [SEP] + o^{i} + [SEP]$. \textsc{QTOnA} utilizes visual information in the form of object names. $y_{text}^{i}$ is obtained using the $q$, $t$ and $o^i$, while $y_{vis}$ is obtained using object names. Similarly in \textsc{QTOfA}, $y_{vis}$ is obtained using object features. Lastly, in \textsc{QTRA}, $y_{vis}$ is obtained using the image features from ResNet.
(5) \textbf{KG facts+Visual+Transcript+Question+Option}: Finally, we also exploit the KG facts for answering the questions. \textsc{QTCFA} extends QTCA by adding KG facts to $B^{i}$ generation. \textsc{QTCOnFA} and \textsc{QTCOnFA} follow the complete approach as illustrated in Fig.~\ref{fig:approach}, the former uses object names while the latter uses object features to obtain $y_{vis}$.

\noindent\textbf{Results}
In Table~\ref{tab:mainResults}, we report the accuracies of individual question types for each experiment. From the table, we make the following observations: (1) In both Longest and Shortest, the performance is very close to random. Although with the longest answers, the accuracy is 8\% more than random accuracy, the random accuracies for all the other cases indicates that there is negligible bias due to the length of the options. (2) BERTSim answers Q3 with 48\% accuracy. Higher performance is also observed in the Q4 type. QA performs close to random. (3) Even with the Question, transcripts and options available, performance gain is absent. Similar to the BERTsim ablation, BERTsim QTA performs better on the Q3 and Q4 types. QTA performs close to random, signifying that questions, options and the transcripts alone are not enough to predict the correct answers. (4) Unsurprisingly, just adding the captions is also not enough to lift accuracies, QTCA performs close to random in all categories. However, incorporating visual information via object names (\textsc{QTOnA}) shows improvements. With minor improvements in other question types, major jumps are observed for the question types Q3, Q4 with 50\% and 32\% increment over the random accuracy. Better performance can also be seen in the Q7, Q8 and Q9 types. With object features (\textsc{QTOfA}) and image features (\textsc{QTRA}) the accuracies still lurk around random, with slight improvements in the Q3 type. (5) Finally, adding the KG facts revamps performance across the board. QTCFA improves over random  by  60\% for Q3 and by 42\% for Q4. It shows a 17\% improvement over random accuracy in Q5, 12\% improvement in Q6, 26\% improvement in Q7, 22\% in Q8 and 20\% in the Q9 type. Adding the object names (\textsc{QTCOnFA}) information along with the captions improves on QTCFA by massive amounts (11.56, 3.54, 33.41 and 23.1 percentage points) for Q5, Q7, Q8 and Q9 resp. Surprisingly, adding the object feature (\textsc{QTCOfFA}) helps but not as much as adding object names, possibly because a simple dense layer is not enough to effectively combine the visual embeddings with the BERT based textual embeddings. The relatively low performance of these methods shows that there is a large scope of work to be done for question answering on news videos.
%, and we aim to investigate further advanced methods as part of future work.

%decreases performance for the majority of cases. This drop possibly indicates that a simple dense layer is not enough to effectively combine the visual embeddings with the BERT based textual embeddings. Although there is a drop in most cases, they show an improvement of 31\% in Q8 and 21\% in Q9 over QTCFA.

\section{Conclusions and Future Work}
We studied the problem of knowledge-aware question answering on videos. We curated a large dataset with 12094 clips and associated them with more than a million questions across 9 different types. The questions have been carefully automatically generated such that each question needs video as well as a KG to be answered. Using  methods which consume signals across different modes (like text, visual and KG), we establish a baseline accuracy across different question types. We observe that there is a lot of room for improvement, thus establishing that our dataset aptly captures the complexity of the video QA task. Video QA has been performed using various kinds of methods like spatio-temporal reasoning~\cite{jiang2020divide}, multi-frame inference using memory-based methods~\cite{fan2019heterogeneous}, question attention-based models~\cite{kim2019progressive} and multimodal attention based methods~\cite{garcia2020knowit}. Since our focus was on building a novel dataset, we presented results using a multimodal baseline method. We plan to adapt these methods for evaluation on \textsc{NewsKVQA} in the future.

% \bibliography{references}

\begin{thebibliography}{10}
\providecommand{\url}[1]{\texttt{#1}}
\providecommand{\urlprefix}{URL }
\providecommand{\doi}[1]{https://doi.org/#1}
\scriptsize
\bibitem{acharya2019tallyqa}
Acharya, M., Kafle, K., Kanan, C.: Tallyqa: Answering complex counting
  questions. In: AAAI. vol.~33, pp. 8076--8084 (2019)

\bibitem{aditya2018explicit}
Aditya, S., Yang, Y., Baral, C.: Explicit reasoning over end-to-end neural
  architectures for visual question answering. In: AAAI. vol.~32 (2018)

\bibitem{antol2015vqa}
Antol, S., Agrawal, A., Lu, J., Mitchell, M., Batra, D., Zitnick, C.L., Parikh,
  D.: Vqa: Visual question answering. In: ICCV. pp. 2425--2433 (2015)

\bibitem{fan2019heterogeneous}
Fan, C., Zhang, X., Zhang, S., Wang, W., Zhang, C., Huang, H.: Heterogeneous
  memory enhanced multimodal attention model for video question answering. In:
  CVPR. pp. 1999--2007 (2019)

\bibitem{garcia2020knowit}
Garcia, N., Otani, M., Chu, C., Nakashima, Y.: Knowit vqa: Answering
  knowledge-based questions about videos. In: AAAI. vol.~34, pp. 10826--10834
  (2020)

\bibitem{guadarrama2013youtube2text}
Guadarrama, S., Krishnamoorthy, N., Malkarnenkar, G., Venugopalan, S., Mooney,
  R., Darrell, T., Saenko, K.: Youtube2text: Recognizing and describing
  arbitrary activities using semantic hierarchies and zero-shot recognition.
  In: ICCV. pp. 2712--2719 (2013)

\bibitem{he2016deep}
He, K., Zhang, X., Ren, S., Sun, J.: Deep residual learning for image
  recognition. In: CVPR. pp. 770--778 (2016)

\bibitem{hu2017learning}
Hu, R., Andreas, J., Rohrbach, M., Darrell, T., Saenko, K.: Learning to reason:
  End-to-end module networks for visual question answering. In: ICCV. pp.
  804--813 (2017)

\bibitem{jang2017tgif}
Jang, Y., Song, Y., Yu, Y., Kim, Y., Kim, G.: Tgif-qa: Toward spatio-temporal
  reasoning in visual question answering. In: CVPR. pp. 2758--2766 (2017)

\bibitem{jiang2020divide}
Jiang, J., Chen, Z., Lin, H., Zhao, X., Gao, Y.: Divide and conquer:
  Question-guided spatio-temporal contextual attention for video question
  answering. In: AAAI. vol.~34, pp. 11101--11108 (2020)

\bibitem{kembhavi2017you}
Kembhavi, A., Seo, M., Schwenk, D., Choi, J., Farhadi, A., Hajishirzi, H.: Are
  you smarter than a sixth grader? textbook question answering for multimodal
  machine comprehension. In: CVPR. pp. 4999--5007 (2017)

\bibitem{kim2019progressive}
Kim, J., Ma, M., Kim, K., Kim, S., Yoo, C.D.: Progressive attention memory
  network for movie story question answering. In: CVPR. pp. 8337--8346 (2019)

\bibitem{kim2017deepstory}
Kim, K.M., Heo, M.O., Choi, S.H., Zhang, B.T.: Deepstory: Video story qa by
  deep embedded memory networks. In: IJCAI. pp. 2016--2022 (2017)

\bibitem{lei2018tvqa}
Lei, J., Yu, L., Bansal, M., Berg, T.L.: Tvqa: Localized, compositional video
  question answering. arXiv:1809.01696  (2018)

\bibitem{lei2020tvqa+}
Lei, J., Yu, L., Berg, T., Bansal, M.: Tvqa+: Spatio-temporal grounding for
  video question answering. In: ACL. pp. 8211--8225 (2020)

\bibitem{li2020oscar}
Li, X., Yin, X., Li, C., Hu, X., Zhang, P., Zhang, L., Wang, L., Hu, H., Dong,
  L., Wei, F., Choi, Y., Gao, J.: Oscar: Object-semantics aligned pre-training
  for vision-language tasks. ECCV  (2020)

\bibitem{lu2018r}
Lu, P., Ji, L., Zhang, W., Duan, N., Zhou, M., Wang, J.: R-vqa: learning visual
  relation facts with semantic attention for visual question answering. In:
  KDD. pp. 1880--1889 (2018)

\bibitem{maharaj2017dataset}
Maharaj, T., Ballas, N., Rohrbach, A., Courville, A., Pal, C.: A dataset and
  exploration of models for understanding video data through fill-in-the-blank
  question-answering. In: CVPR. pp. 6884--6893 (2017)

\bibitem{marino2019ok}
Marino, K., Rastegari, M., Farhadi, A., Mottaghi, R.: Ok-vqa: A visual question
  answering benchmark requiring external knowledge. In: CVPR. pp. 3195--3204
  (2019)

\bibitem{mun2017marioqa}
Mun, J., Hongsuck~Seo, P., Jung, I., Han, B.: Marioqa: Answering questions by
  watching gameplay videos. In: ICCV. pp. 2867--2875 (2017)

\bibitem{reimers2019sentence}
Reimers, N., Gurevych, I.: Sentence-bert: Sentence embeddings using siamese
  bert-networks. arXiv:1908.10084  (2019)

\bibitem{ren2015faster}
Ren, S., He, K., Girshick, R.B., Sun, J.: Faster r-cnn: Towards real-time
  object detection with region proposal networks. In: NIPS. p. 91–99 (2015)

\bibitem{shah2019kvqa}
Shah, S., Mishra, A., Yadati, N., Talukdar, P.P.: Kvqa: Knowledge-aware visual
  question answering. In: AAAI. vol.~33, pp. 8876--8884 (2019)

\bibitem{singh2019towards}
Singh, A., Natarajan, V., Shah, M., Jiang, Y., Chen, X., Batra, D., Parikh, D.,
  Rohrbach, M.: Towards vqa models that can read. In: CVPR. pp. 8317--8326
  (2019)

\bibitem{tapaswi2016movieqa}
Tapaswi, M., Zhu, Y., Stiefelhagen, R., Torralba, A., Urtasun, R., Fidler, S.:
  Movieqa: Understanding stories in movies through question-answering. In:
  CVPR. pp. 4631--4640 (2016)

\bibitem{wang2017explicit}
Wang, P., Wu, Q., Shen, C., Dick, A., van~den Hengel, A.: Explicit
  knowledge-based reasoning for visual question answering. In: IJCAI. pp.
  1290--1296 (2017)

\bibitem{wang2017fvqa}
Wang, P., Wu, Q., Shen, C., Dick, A., Van Den~Hengel, A.: Fvqa: Fact-based
  visual question answering. TPAMI  \textbf{40}(10),  2413--2427 (2017)

\bibitem{wu2016ask}
Wu, Q., Wang, P., Shen, C., Dick, A., Van Den~Hengel, A.: Ask me anything:
  Free-form visual question answering based on knowledge from external sources.
  In: CVPR. pp. 4622--4630 (2016)

\bibitem{xu2017video}
Xu, D., Zhao, Z., Xiao, J., Wu, F., Zhang, H., He, X., Zhuang, Y.: Video
  question answering via gradually refined attention over appearance and
  motion. In: ACM MM. pp. 1645--1653 (2017)

\bibitem{yang2016stacked}
Yang, Z., He, X., Gao, J., Deng, L., Smola, A.: Stacked attention networks for
  image question answering. In: CVPR. pp. 21--29 (2016)

\bibitem{yu2019activitynet}
Yu, Z., Xu, D., Yu, J., Yu, T., Zhao, Z., Zhuang, Y., Tao, D.: Activitynet-qa:
  A dataset for understanding complex web videos via question answering. In:
  AAAI. vol.~33, pp. 9127--9134 (2019)

\bibitem{zadeh2019social}
Zadeh, A., Chan, M., Liang, P.P., Tong, E., Morency, L.P.: Social-iq: A
  question answering benchmark for artificial social intelligence. In: CVPR.
  pp. 8807--8817 (2019)

\bibitem{zellers2019recognition}
Zellers, R., Bisk, Y., Farhadi, A., Choi, Y.: From recognition to cognition:
  Visual commonsense reasoning. In: CVPR. pp. 6720--6731 (2019)

\bibitem{zeng2017leveraging}
Zeng, K.H., Chen, T.H., Chuang, C.Y., Liao, Y.H., Niebles, J.C., Sun, M.:
  Leveraging video descriptions to learn video question answering. In: AAAI.
  vol.~31 (2017)

\bibitem{zhu2015building}
Zhu, Y., Zhang, C., R{\'e}, C., Fei-Fei, L.: Building a large-scale multimodal
  knowledge base system for answering visual queries. arXiv:1507.05670  (2015)

\end{thebibliography}

\end{document}